\documentclass{article}

    \PassOptionsToPackage{numbers, compress}{natbib}
    \usepackage[preprint]{neurips_2020}

\usepackage[utf8]{inputenc} % allow utf-8 input
\usepackage[T1]{fontenc}    % use 8-bit T1 fonts
\usepackage{hyperref}       % hyperlinks
\usepackage{url}            % simple URL typesetting
\usepackage{booktabs}       % professional-quality tables
\usepackage{amsfonts}       % blackboard math symbols
\usepackage{nicefrac}       % compact symbols for 1/2, etc.
\usepackage{microtype}      % microtypography

\usepackage{natbib}
\usepackage{graphicx}
\usepackage{subfigure}
\usepackage{amssymb}
\usepackage{todonotes}
\usepackage{amsmath}
\usepackage{makecell}
\usepackage{tikz}
\usepackage[outline]{contour}
\usepackage{nccmath}
\usepackage{caption}     
\usepackage{multirow}
\usepackage{fix-cm}
\usepackage{siunitx}
\usepackage{multicol}

\title{Data augmentation with Mobius transformations}
\author{Sharon Zhou$^\dagger$, Jiequan Zhang$^\dagger$, Hang Jiang$^\dagger$, Torbj\"orn Lundh$^\ddagger$, Andrew Y. Ng$^\dagger$ \\
$^\dagger$Stanford University, $^\ddagger$Chalmers University of Technology \\
$^\dagger$\texttt{\{sharonz, jiequanz, hjian42, ang\}@cs.stanford.edu} \\
$^\ddagger$\texttt{torbjorn.lundh@chalmers.se} \\
}

\begin{document}

\maketitle

\begin{abstract}
% Abstracts must be a single paragraph, ideally between 4--6 sentences long.
% Data augmentation has led to substantial improvements in the performance and generalization of deep models, and remain a highly adaptable method to evolving model architectures and varying amounts of data---in particular, extremely scarce amounts of available training data. In this paper, we present a novel method of applying M\"obius transformations to augment input images during training. M\"obius transformations are bijective conformal maps that generalize image translation to operate over complex inversion in pixel space. As a result, M\"obius transformations can operate on the sample level and preserve data labels. We also apply M\"obius to augment text embeddings for text classification and suggest methods for inclusion beyond the image domain. We show empirically that the inclusion of M\"obius transformations during training enables improved generalization over prior sample-level data augmentation techniques, such as cutout and standard crop-and-flip transformations on images and the BERT baseline on text.
% , most notably in low data regimes.
% pixel space -> feature space
Data augmentation has led to substantial improvements in the performance and generalization of deep models, and remain a highly adaptable method to evolving model architectures and varying amounts of data---in particular, extremely scarce amounts of available training data. In this paper, we present a novel method of applying M\"obius transformations to augment input images during training. M\"obius transformations are bijective conformal maps that generalize image translation to operate over complex inversion in pixel space. As a result, M\"obius transformations can operate on the sample level and preserve data labels. We show that the inclusion of M\"obius transformations during training enables improved generalization over prior sample-level data augmentation techniques such as cutout and standard crop-and-flip transformations, most notably in low data regimes.
\end{abstract}
% 8 page limit

\begin{figure}[ht]
\begin{centering}
  \includegraphics[width=.9\columnwidth]{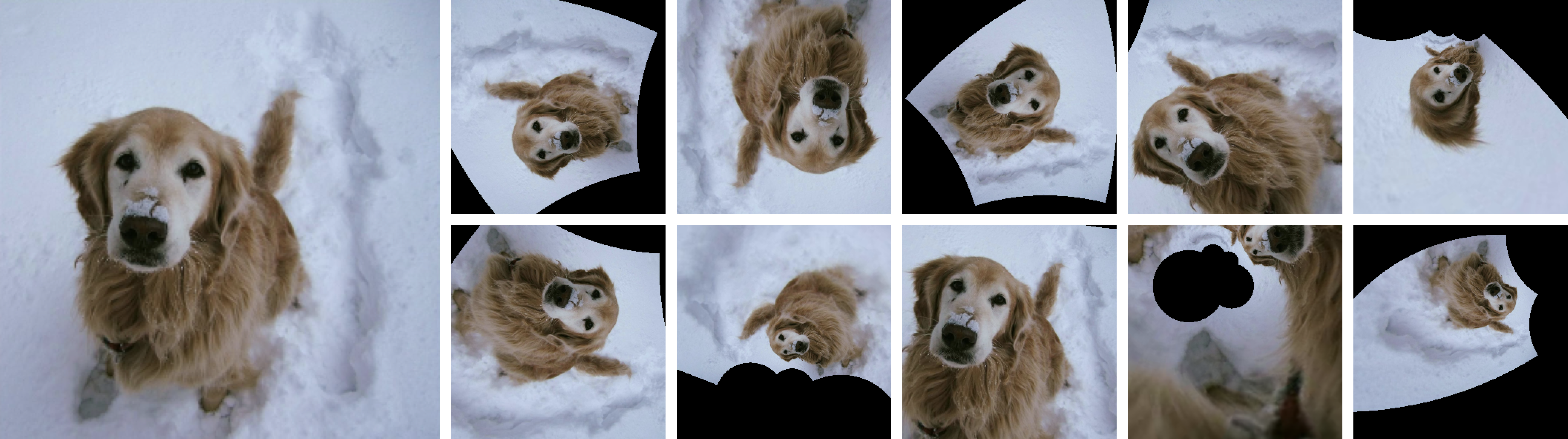}
  \caption{Examples of M\"obius transformations (original on left), resulting in variations in perspective, orientation, and scale, while still preserving local angles and anharmonic ratios.}
  \label{fig:pull}
\end{centering}
\end{figure}

\section{Introduction}
Data augmentation has significantly improved the generalization of deep neural networks on a variety of image tasks, including image classification~\cite{perez2017effectiveness, devries2017improved}, object detection~\cite{zhong2017random, guo2015deep}, and instance segmentation~\cite{wang2018understanding}. Prior work has shown that data augmentation on its own can perform better than, or on par with, highly regularized models using other regularization techniques such as dropout~\cite{hernandez2018data}. This effectiveness is especially prominent in low data regimes, where models often fail to capture the full variance of the data in the training set~\cite{zhang2019dada}. 

Many data augmentation techniques rely on priors that are present in the natural world. Standard operations, such as translation, crop, and rotation, in addition to more recent methods, such as cutout~\cite{devries2017improved}, have improved generalization by encouraging equivariance to the transformation. For example, an image of a horse remains a horse in its vertical reflection or with its body partially occluded. As a result, these transformations are able to preserve the original label on the augmented sample, enabling straightforward and easy incorporation into the growing number of data augmentation algorithms for both fully supervised and semi-supervised learning. In a nutshell, these sample-level methods have been not only effective, but also interpretable, easy to implement, and flexible to incorporate.

Following the success of these methods, we focus this paper on augmentations that exploit natural patterns to preserve labels after transformation and that operate on the sample level. These transformations easily complement other methods, thus leveraged in a wide variety of data augmentation algorithms. In contrast, multi-sample augmentations, which have had comparably strong empirical results~\cite{zhang2017mixup}, unfortunately connect less clearly to natural priors that would support equivariance to the augmentation. While performant on their own, these methods have had less success with integration into data augmentation algorithms and policies~\cite{cubuk2019autoaugment, cubuk2019randaugment, xie2019unsupervised}, except for those tailored to them~\cite{berthelot2019mixmatch}.

In this paper, we propose a novel data augmentation technique, inspired by biological patterns, using bijective conformal maps known as \textit{M{\"o}bius transformations}. M{\"o}bius transformations perform complex inversion in pixel space, extending standard translation to include divisibility. These transformations enable perspective projection---or transforming perceived distance of objects in an image---and are found naturally in the anatomy and biology of humans and other animals. 

We define a class of $\mathcal{M}$-admissible M{\"o}bius transformations that preserves image-level labels by minimizing local distortions in an image. We show empirically that the inclusion of $\mathcal{M}$-admissible M{\"o}bius transformations can improve performance on the CIFAR-10, CIFAR-100, and Tiny ImageNet benchmarks over prior sample-level data augmentation techniques, such as cutout~\cite{devries2017improved} and standard crop-and-flip baselines. We additionally show that M\"obius transformations successfully complement other transformations. 
% M{\"o}bius transformations are especially useful in low data settings particularly when applied in small quantities. Finally, w
%  during training to improve the generalization of deep models

Our key contributions can be summarized as follows:
\begin{itemize}
    \item \textbf{Method:} We introduce a class of $\mathcal{M}$-admissible M{\"o}bius transformations for data augmentation in training neural networks. This M\"obius class allows for a wide range of sample-level mappings that preserve local angles and can be found in the anatomy of animals.
    \item \textbf{Performance:} Empirically, the inclusion of M{\"o}bius data augmentation improves model generalization over prior methods that use sample-level augmentation techniques, such as cutout~\cite{devries2017improved} and standard crop-and-flip transformations. We also show that M\"obius transformations, which have been studied and examined in the anatomy and biology of animals, consistently improves on animate classes over inanimate classes.
    \item \textbf{Low data:} M{\"o}bius is especially effective in low data settings, where the data quantity is on the order of hundreds of samples per class. 
    % \item \textbf{Complementary use:} M\"obius data augmentation complements other methods, e.g. cutout. Implemented together, they can significantly outperform either alone.
    % \item \textbf{Animate analysis:} We find that M\"obius transformations, which have been studied and examined in the anatomy and biology of animals, consistently and significantly improves on animate classes over inanimate ones across datasets. 
\end{itemize}

\section{M{\"o}bius transformations}
M\"obius transformations are bijective conformal mappings that operate over complex inversion and preserve local angles. They are also known as bilinear or linear fractional transformations. We discuss their biological and perceptual underpinnings, and follow with a formal definition. Finally, we describe their application to data augmentation to improve generalization of convolutional neural networks on image classification tasks.

\subsection{Motivation}
M\"obius transformations have been studied in biology as 2D projections of specimens---such as humans, fungi, and fish---from their 3D configurations~\cite{thompson1942growth, petukhov1989non, Lundh2011Cross-Development}. Mathematically, most of these examples leverage Liouville's theorem~\cite{liouville1850extension}, which states that smooth conformal mappings are M\"obius transformations on a domain of $\mathbb{R}^n$ where $n>2$. These biological patterns motivate our application of M\"obius transformations to natural images, particularly those that include the relevant species.

Beyond biological underpinnings, M\"obius transformations preserve the \textit{anharmonic ratio}~\cite{ahlfors1989mobius, needham1998visual}, or the extent to which four collinear points on a projective line deviate from the harmonic ratio.\footnote{The anharmonic ratio, also denoted \textit{cross-ratio}, stems from projective geometry and has been studied in biology with respect to M\"obius transformations~\cite{petukhov1989non, Lundh2011Cross-Development}.} This invariance is a property that M\"obius transformations share with projective transformations, which are used widely in metrology~\cite{criminisi2000single}. In the context of transforming natural images, such a transformation can be particularly useful for perspective projection. That is, an image can be transformed to an alternate perceived distance. This effect is visually apparent across examples in Figure~\ref{fig:pull}. 

\subsection{Definition}
Existing data augmentation techniques for image data belong to the class of affine mappings, i.e. the group of translation, scaling, and rotation, which can be generally described using a complex function $z\to az+b$, where the variable $z$ and the two parameters $a$ and $b$ are complex numbers. M\"obius transformations represent the next level of abstraction by introducing division to the operation~\cite{lie1893theorie,petukhov1989non}. The group of M\"obius transformations can be described as all functions $f$ from $ \mathbb{C} \rightarrow \mathbb{C}$ with the form
\begin{equation}
f(z) = \frac{az+b}{cz+d},
\label{eq:def}
\end{equation}
where $a,b,c,d \in \mathbb{C}$ such that $ad-bd\neq 0$. As a result, the set of all M\"obius transformations is a superset of several basic transformations, including translation, rotation, inversion, and an even number of reflections over lines. 

One method for programatically implementing such a transformation with complex values $a,b,c,d$ in Eq.~(1) is to use the fact that there exists a unique M\"obius transformation sending any three points to any three other points in the extended complex plane~(\citealp[p. 150]{needham1998visual}). That is, equivalent to specifying $a,b,c$ and $d$ directly in~(1), we can define three separate points $z_1, z_2, z_3 \in \mathbb{C}$ in the image and then select three separate target points $w_1, w_2, w_3 \in \mathbb{C}$, to which those initial points will be mapped in the resulting transformation. From these two sets of points, we can then compute the values of the transformation using the knowledge that anharmonic ratios---adding the points $z_i$ and $w_i$ where $i=\{1,2,3\}$ completes the two quartets---are Möbius invariant~(\citealp[p. 154]{needham1998visual}), resulting in the following equality:

\begin{equation} \frac{(w-w_1)(w_2-w_3)}{(w-w_3)(w_2-w_1)} = \frac{(  z-  z_1)(  z_2-  z_3)}{(  z-  z_3)(  z_2-  z_1)}. \label{eq:cr}
\end{equation}

We can rearrange this expression by solving for $w$:

\[\frac{w-w_1}{w-w_3}=\frac{(  z-  z_1)(  z_2-  z_3)(w_2-w_1)}{(  z-  z_3)(  z_2-  z_1)(w_2-w_3)}\]

\[ w=\frac{Aw_3 - w_1}{A-1}\]

where $A = \frac{(  z-  z_1)(  z_2-  z_3)(w_2-w_1)}{(  z-  z_3)(  z_2-  z_1)(w_2-w_3)}$. This final expression for $w$ is in the form of Eq.~\eqref{eq:def}:
\begin{equation*}
    f(z)= w =\frac{Aw_3 - w_1}{A-1}=\frac{az+b}{cz+d},
    \label{eq:equiv}
\end{equation*}
% \[w=\frac{az+b}{cz+d}\]
from which we can compute the following values for $a,b,c, $ and $d$ using basic algebraic operations: 
\begin{equation*}
a =  w_1 w_2 z_1 - w_1 w_3 z_1 - w_1 w_2 z_2 + w_2 w_3 z_2 + w_1 w_3 z_3 - w_2 w_3 z_3,
 \end{equation*}

\begin{equation*}
b= w_1 w_3 z_1 z_2 - w_2 w_3 z_1 z_2 - w_1 w_2 z_1 z_3 + w_2 w_3 z_1 z_3 + w_1 w_2 z_2 z_3 - w_1 w_3 z_2 z_3,  \end{equation*}

\begin{equation*}
c= w_2 z_1 - w_3 z_1 - w_1 z_2 + w_3 z_2 + w_1 z_3 - w_2 z_3,  
\end{equation*}

\begin{equation*}
d=w_1 z_1 z_2 - w_2 z_1 z_2 - w_1 z_1 z_3 + w_3 z_1 z_3 + w_2 z_2 z_3 - w_3 z_2 z_3.
\end{equation*}

  Alternatively, by solving Eq.~\eqref{eq:cr} using linear algebra, i.e. evaluating a determinant from this construction using the Laplace expansion, one can elegantly express these algebraic expressions above as determinants: 

\begin{equation*}
a =  
    \begin{vmatrix}
    z_1w_1 & w_1 & 1 \\
    z_2w_2 & w_2 & 1 \\
    z_3w_3 & w_3 & 1
    \end{vmatrix}, \; \; \; \;
b = 
    \begin{vmatrix}
    z_1w_1 &   z_1   &   w_1   \\
      z_2w_2   &   z_2   &   w_2   \\
      z_3w_3   &   z_3   &   w_3 
    \end{vmatrix}, \; \; \; \;
c = 
    \begin{vmatrix}
      z_1   &   w_1   & 1 \\
      z_2   &   w_2   & 1 \\
      z_3   &   w_3   & 1
    \end{vmatrix}, \; \; \; \; 
d = 
    \begin{vmatrix}
      z_1w_1   &   z_1   & 1 \\
      z_2w_2   &   z_2   & 1 \\
      z_3w_3   &   z_3   & 1
    \end{vmatrix}.
\end{equation*}

This pointwise method is used in our work to construct valid image augmentations using M\"obius transformations. Ultimately, this method can be leveraged to define specific types of M\"obius transformations programmatically for needs within and beyond data augmentation.

\subsubsection{Equivalent framing: circle reflection}
We introduce an equivalent formulation of M\"obius transformations on images in $\mathbb{R}^2$. The goal of this section is to lend intuition on constraints that we apply to M\"obius data augmentation in Section~\ref{sec:dataaug} that follows. 

M\"obius mappings in the plane can also be defined as the set of transformations with an even number of reflections over circles and lines (i.e. circles with infinite radii) on the plane. A reflection, or inversion, in the unit circle is the complex transformation (\citealp[pp. 124]{needham1998visual}):
\begin{equation*}
    z \in \mathbb{C} \rightarrow \frac{z}{|z|^2}.
    \label{eq:circle}
\end{equation*} 
%Since $|z|^2=z\cdot \bar{z}$, then $\frac{z}{|z|^2} =\frac{1}{\bar{z}}$ (\citealp[pp. 124]{needham1998visual}). 

Thus, a M\"obius transformation on an image is simply a reflection over the unit circle, with pixels inside of the circle projected outwards and pixels on the outside projected inwards. As such, M\"obius transformations often reflect a different amount of pixels inwards as opposed to outwards, and this imbalance enables the scale distortions seen in Figure~\ref{fig:pull}. Note that a circular shape can be left as an artifact after the transformation, if the reflection occurs at an edge without any pixels to project inwards. 

\section{Class of $\mathcal{M}$-admissible M\"obius transformations}
\label{sec:dataaug}
In order to use M{\"o}bius transformations for data augmentation, we need to constrain the set of possible transformations. When taken to the limit, M\"obius transformations do not necessarily preserve the image label. This is similar to constraining translation in order to ensure that pixels remain afterwards, or to keeping cutout to lengths judiciously less than the size of the image so that it is not fully occluded. Because M{\"o}bius transformations inherently reflect more pixels in one direction (into or out of the circle), we will often see two main effects:~(1) incongruent sizes of the output from the initial input and~(2) gaps between pixels in the result transformation, sometimes significant depending on the location of the circle. E.g., if the circle is placed at the edge of the image, there is little to project from the edge inwards. To address both of these effects, we enforce equal sizing after the transformation and cubic spline interpolation during reflection to fill gaps.

In tandem, we introduce a class of $\mathcal{M}$-admissible M\"obius transformations that control the local distortion in an image, in order to avoid explosive and implosive mappings, by bounding the modulus of the derivative above and below. If we view each pixel as a circle to be mapped at another circle (pixel) and use the analogue of the $f^{\#}$-function, as defined in an authoritative text on M\"obius~\cite{dubejko1995circle}, to the modulus of the derivative of the M\"obius transformation, $|f'|$, we can bound it above and below by two constants, or for simplicity, one real constant $M>1$ such that:
\begin{equation} \frac{1}{M} < |f'| < M.\label{eq:Mcondition} \end{equation}
As an approximation, we will only check this condition for only five points on an image of size $[0,p]\times[0,pi]$: the four inverse images of the corner points in the square $[0,p]\times[0,pi]$ and the center point: $\frac{1}{2}p(1+i)$. Furthermore, in order to only consider transformations that will keep enough information from the original picture, we add the condition that the pre-image of the center point, $f^{-1}(\frac{1}{2}p(1+i))$, should be inside the centered circle half-way to the sides, i.e.
\begin{equation} |f^{-1}(\frac{1}{2}p(1+i))-\frac{1}{2}p(1+i)|<\frac{p}{4}].\label{eq:centerpoint} 
\end{equation}
To give more concrete and computable conditions \eqref{eq:Mcondition} and  \eqref{eq:centerpoint}, we start with a general M\"obius transformation from definition \eqref{eq:def} with the condition that $ad\neq bc$ to obtain the inverse $f^{-1}$ as
\begin{equation} f^{-1}(z)=-\frac{d z-b}{c z-a},\label{eq:finv}
\end{equation}
For the condition in \eqref{eq:centerpoint}, we compute \eqref{eq:finv}
\[f^{-1}(\frac{1}{2}p(1+i))=\frac{\frac{1}{2}p(1+i)d -b}{a-\frac{1}{2}p(1+i)c}.\]
For the condition \eqref{eq:Mcondition}, we compute the derivative $f'$
\begin{equation} f'(z)=\frac{a}{c z+d}-\frac{c (a z+b)}{(c z+d)^2}.\label{eq:fprim} \end{equation}

Combining \eqref{eq:finv} and \eqref{eq:fprim} and simplifying, we obtain the following simple expression:
\begin{equation} 
f'(f^{-1}(z))=\frac{(a-c z)^2}{a d -b c}.
\label{eq:combinedfunctions}
\end{equation}

By using \eqref{eq:combinedfunctions} and \eqref{eq:finv}, we can give a reformulation of the conditions in \eqref{eq:Mcondition} and \eqref{eq:centerpoint} to define a subclass of all M\"obius transformations,
\[f(z) = \frac{az+b}{cz+d}, \mbox{ where } ad-bc \neq 0,\] which we call the class of \textbf{$\mathcal{M}$-admissible M\"obius transformations} as long as the function $f$ fulfills the following list of inequalities by checking the points $0,p,pi,p(1+i),\frac{1}{2}p(1+i)$: 
% \begin{equation}
% \begin{array}{rcl}
% \frac{1}{M} < &\frac{|a|^2}{|ad-bc|} &< M, \\ \\
% \frac{1}{M} < &\frac{|a-pc|^2}{|ad-bc|} &< M,\\ \\
% \frac{1}{M} < &\frac{|pc+ai|^2}{|ad-bc|} &< M, \\ \\
% \frac{1}{M} < &\frac{|a-p(1+i)c|^2}{|ad-bc|} &< M, \\ \\
% \frac{1}{M} < &\frac{|a-\frac{1}{2}p(1+i)c|^2}{|ad-bc|} &< M,\\ \\
% \big|\frac{\frac{1}{2}p(1+i)d -b}{a-\frac{1}{2}p(1+i)c} \!\!\!\!& - \; \frac{1}{2}p(1+i)\big| &< \frac{p}{4}. 
% \end{array}
% \label{eq:Madmissable}
% \end{equation}

\begin{align}
\frac{1}{M} < &\frac{|a|^2}{|ad-bc|} < M, & \nonumber &\frac{1}{M} < \frac{|a-pc|^2}{|ad-bc|} < M,\\ \nonumber \\ \nonumber
\frac{1}{M} < &\frac{|pc+ai|^2}{|ad-bc|} < M, & &\frac{1}{M} < \frac{|a-p(1+i)c|^2}{|ad-bc|} < M,\\ \nonumber \\
\frac{1}{M} < &\frac{|a-\frac{1}{2}p(1+i)c|^2}{|ad-bc|} < M, & \big|&\frac{\frac{1}{2}p(1+i)d -b}{a-\frac{1}{2}p(1+i)c} \!\!\!\! - \; \frac{1}{2}p(1+i)\big| < \frac{p}{4}. \nonumber
\end{align}

% \begin{multicols}{2}
%   \begin{equation}
% \frac{1}{M} < \frac{|a|^2}{|ad-bc|} < M, 
%   \end{equation}
%   \begin{equation}
% \frac{1}{M} < \frac{|a-pc|^2}{|ad-bc|} < M,
%   \end{equation}\break
%   \begin{equation}
% \frac{1}{M} < \frac{|pc+ai|^2}{|ad-bc|} < M, 
%   \end{equation}
%   \begin{equation}
% \frac{1}{M} < \frac{|a-p(1+i)c|^2}{|ad-bc|} < M, 
%   \end{equation}\break
%     \begin{equation}
%   \frac{1}{M} < \frac{|a-\frac{1}{2}p(1+i)c|^2}{|ad-bc|} < M, 
%   \end{equation}
%   \begin{equation}
% \big|\frac{\frac{1}{2}p(1+i)d -b}{a-\frac{1}{2}p(1+i)c} \!\!\!\! - \; \frac{1}{2}p(1+i)\big| < \frac{p}{4}. 
%   \end{equation}

% \end{multicols}

Sampling from the $\mathcal{M}$-admissible class, we can incorporate label-preserving M{\"o}bius transformations into classical data augmentation methods of the form $(x,y)=(f(x),y)$, where $f$ here is a M{\"o}bius transformation on an image $x$, preserving label $y$.

\section{Related work}
A large number of data augmentation techniques have recently emerged for effectively regularizing neural networks, including both sample-level augmentations, such as ours, as well as multi-sample augmentations that mix multiple images. We discuss these, as well as data augmentation algorithms that leverage multiple augmentations. Finally, we examine ways in which M\"obius transformations have been applied to deep learning. To our knowledge, this is the first work using M{\"o}bius transformations for data augmentation in deep neural networks.

\subsection{Data augmentation}
\textbf{Sample-level augmentation}. M{\"o}bius transformations generalize standard translation to include inversion as an operation under conformity, demonstrating outputs that appear to have gone through crop, rotation, and/or scaling, while preserving local angles from the original image. We recognize that the list of image transformations is extensive: crop, rotation, warp, skew, shear, distortion, Gaussian noise, among many others. Additional sample-level data augmentation methods use occlusion such as cutout~\cite{devries2017improved} and random erasing~\cite{zhong2017random}, which apply random binary masks across image regions.

\textbf{Multi-sample augmentation}. Data augmentation on images also consists of operations applied to multiple input images. In such cases, original labels are often mixed. For example, MixUp~\cite{zhang2017mixup} performs a weighted average of two images (over pixels) and their corresponding labels in varying proportions to perform soft multi-label classification. Between-class learning~\cite{tokozume2018between} and SamplePairing~\cite{inoue2018data} are similar techniques, though the latter differs in using a single label. Comparably, RICAP~\cite{takahashi2019data}, VH-Mixup and VH-BC+~\cite{summers2019improved} form composites of several images into one. While these methods have performed well, we focus this paper on comparisons to sample-level augmentations that preserve original labels and that can be more readily incorporated into data augmentation policies.

\textbf{Algorithms and policies for data augmentation}. Various strategies have emerged to incorporate multiple data augmentation techniques for improved performance. AutoAugment~\cite{cubuk2019autoaugment}, Adatransform~\cite{tang2019adatransform}, RandAugment~\cite{cubuk2019randaugment}, and Population Based Augmentation~\cite{ho2019population} offer ways to select optimal transformations (and their intensities) during training. In semi-supervised learning, unsupervised data augmentation~\cite{xie2019unsupervised}, MixMatch~\cite{berthelot2019mixmatch}, and FixMatch~\cite{sohn2020fixmatch} have shown to effectively incorporate unlabeled data by exploiting label preservation and consistency training. Tanda~\cite{ratner2017learning} composes sequences of augmentation methods, such as crop followed by cutout then flip, that are tuned to a certain domain. DADA~\cite{zhang2019dada} frames data augmentation as an adversarial learning problem and applies this method in low data settings. We do not test all of these augmentation schemes: our results suggest that M{\"o}bius could add value as an addition to the search space of possible augmentations, e.g. in AutoAugment, or as a transformation that helps enforce consistency between original and augmented data, e.g. in unsupervised data augmentation.

\subsection{M{\"o}bius transformations in deep learning}
M{\"o}bius transformations have been previously studied across a handful of topics in deep learning. Specifically, they have been used as building blocks in new activation functions~\cite{ozdemir2011complex} and as operations in hidden layers~\cite{zammit2019deep}. Coupled with the theory of gyrovector spaces, M{\"o}bius transformations have inspired hyperbolic neural networks~\cite{ganea2018hyperbolic}. They also play an important component in deep fuzzy neural networks for approximating the Choquet integral~\cite{islam2019enabling}. Finally, model activations and input-output relationships have been theoretically related to M{\"o}bius transformations~\cite{mandic2000use}. While prior work has primarily leveraged them for architectural contributions, our work is the first to our knowledge to introduce M{\"o}bius transformations for data augmentation and their empirical success on image classification benchmarks.
\newcolumntype{P}[1]{>{\centering\arraybackslash}p{#1}}

\begin{table*}[t]
\centering
\begin{tabular}{P{3.2cm}P{2.4cm}P{2cm}P{1.7cm}P{2.5cm}}
\Xhline{2\arrayrulewidth}
\textbf{Augmentation Method}             & \textbf{Dataset}  & \textbf{Images Per Class} & \textbf{\# Training Images} & \textbf{Accuracy} \\
\hline
Crop-and-flip                         & CIFAR-10  & 5000   & 50k   & 96.47\%   $\pm 0.04$    \\
$\text{Cutout}_{l=16}$          & CIFAR-10  & 5000   & 50k   & \textbf{97.13}\%   $\pm \textbf{0.03}$      \\
M\"obius                          & CIFAR-10  & 5000   & 50k   & 96.67\%   $\pm 0.13$    \\
M\"obius + $\text{Cutout}_{l=16}$ & CIFAR-10  & 5000   & 50k   & \textbf{97.10}\%   $\pm \textbf{0.16}$  \\
\hline
Crop-and-flip                         & CIFAR-100 & 600    & 50k   & 81.91\%   $\pm 0.20$      \\
$\text{Cutout}_{l=8}$           & CIFAR-100 & 600    & 50k   & 82.35\%   $\pm 0.19$     \\
M\"obius                          & CIFAR-100 & 600    & 50k   & 82.48\% $\pm 0.38$      \\
M\"obius + $\text{Cutout}_{l=8}$  & CIFAR-100 & 600    & 50k   & \textbf{82.67\%}   $\pm \textbf{0.21}$  \\

\hline
% Crop-and-flip                         & Tiny ImageNet  & 500    & 100k    & 52.64\%    $\pm 0.95$     \\
% $\text{Cutout}_{l=16}$          & Tiny ImageNet  & 500    & 100k    & 54.35\%   $\pm 0.02$     \\
% M\"obius                          & Tiny ImageNet  & 500    & 100k    & \textbf{56.64\%}   $\pm \textbf{0.60}$ \\
% M\"obius + $\text{Cutout}_{l=16}$ & Tiny ImageNet  & 500    & 100k    & 56.37\%  $\pm 0.31$     \\
Crop-and-flip                         & Tiny ImageNet  & 500    & 100k    & 68.40\%    $\pm 0.33$     \\
$\text{Cutout}_{l=16}$          & Tiny ImageNet  & 500    & 100k    & 68.64\%   $\pm 0.40$     \\
M\"obius                          & Tiny ImageNet  & 500    & 100k    & 69.04\%   $\pm 0.42$ \\
M\"obius + $\text{Cutout}_{l=16}$ & Tiny ImageNet  & 500    & 100k    & \textbf{69.51\%}  $\pm \textbf{0.25}$     \\
\Xhline{2\arrayrulewidth}
\end{tabular}
\newline\newline
\caption{Experimental results on several dataset settings, including CIFAR-10, CIFAR-100, and Tiny ImageNet. M\"obius performs best empirically on low data settings, such as Tiny ImageNet and CIFAR-100, where the number of images per class is on the order of hundreds. }
% Rows are sorted by ascending average performance within each dataset setting to observe rank.
%  On CIFAR-10, cutout exceeds M\"obius alone, and both outperform the crop-and-flip baseline.
\label{tab:main}
\end{table*}

\section{Experiments}
We experiment on CIFAR-10, CIFAR-100, and Tiny ImageNet. The CIFAR-10 and CIFAR-100 image classification benchmarks use standard data splits of 50k training and 10k test~\cite{krizhevsky2009learning}. CIFAR-10 has 10 classes with 10k images per class, while CIFAR-100 has 100 classes with 500 images per class, in their training sets. Finally, we experiment on Tiny ImageNet~\cite{wu2017tiny}, a subset of ImageNet that still includes ImageNet's variability and higher resolution imagery, while needing fewer resources and infrastructure than running the full ImageNet dataset. The training set constitutes 100k images across 200 classes, and the test set contains 10k images.
% We additionally evaluate reduced CIFAR-10, a common low data regime for which only a subset of the training data---4k images with 400 images per class---is used. We do so in a fully supervised manner, without using unlabeled data. 
%  (a reason similar to the mini-ImageNet dataset in meta-learning literature~\cite{vinyals2016matching})

Thus, we explore three dataset settings: (1)~CIFAR-10, (2)~CIFAR-100, and (3)~Tiny ImageNet. The goal of these experiments is to assess the fundamental concept of including M\"obius data augmentation across data settings. %during training in both regular and low data settings. 
% For examples of M\"obius transformations across different classes and, see Figure~\ref{fig:tiny}.
% Thus, we explore four dataset settings: (1)~CIFAR-10, (2)~CIFAR-100, (3)~reduced CIFAR-10, and (4)~Tiny ImageNet. The goal of these experiments is to assess the fundamental concept of including M\"obius data augmentation during training in both regular and low data settings.
% Already quite low-sample per class, CIFAR-100 is not readily reduced in most prior work. 

\subsection{Evaluation of benchmarks}
\label{sec:experiments-main}
Following prior work on introducing novel data augmentation methods~\cite{devries2017improved, cubuk2019autoaugment}, we use standard crop-and-flip transformations as the baseline across all experimental conditions. We design our experiments to both compare to, and complement, cutout~\cite{devries2017improved}, the previous state-of-the-art image transformation that operates on the sample level, preserves labels, and thus has been easy to incorporate into data augmentation policies. Cutout and standard crop-and-flip also remain the default augmentation choices in recent work~\cite{cubuk2019autoaugment}. Thus, we compare the following conditions: (1)~baseline with only crop and flip, (2)~cutout, (3)~M\"obius, and (4)~M\"obius with cutout. Note that all conditions incorporate crop and flip transformations, following the original cutout paper~\cite{devries2017improved}. Because all augmentation techniques are sample-level and preserve labels, they are complementary and can be layered on each other. We further explore these effects by combining M\"obius with cutout in our experiments.

We draw from prior work on cutout~\cite{devries2017improved} to set the training procedure across all experiments. For cutout, we tune hyperparameter values for each dataset based on prior work~\cite{devries2017improved}.  Note that we select this setup to optimize for cutout and compare directly to their work; it is possible that better hyperparameters exist for M\"obius. We incorporate M\"obius augmentation twenty percent of the time in all experiments, and show further improvements varying its inclusion on different data settings in the Appendix. On the Tiny ImageNet dataset, for which cutout did not present a baseline, we use a standard residual network, average across three runs, and train on 2 NVIDIA Tesla V100 GPUs. Finally, we compute significance using independent t-tests between sample performances of pairwise conditions across runs. 

% Differences are from balancing computational resources with computational efficiency to work on the larger dataset.
% On CIFAR-10 and CIFAR-100, we use a standard wide residual network~\cite{zagoruyko2016wide}, evaluate performance across five runs, and train models using 4 NVIDIA GeForce GTX 1070 GPUs.
% M\"obius does not require any fine-tuning across datasets. Cutout has hyperparameter values tuned for each dataset~\cite{devries2017improved}: a length of 16 on CIFAR-10 ($\text{cutout}_{l=16}$) and a length of 8 on CIFAR-100 and Tiny ImageNet ($\text{cutout}_{l=8}$).

% As shown in Table~\ref{tab:main}, these experiments highlight a couple key observations:
% \begin{itemize}
%     \item Empirically, M\"obius demonstrates significant improvements over cutout and standard crop-and-flip on CIFAR-100 and Tiny ImageNet, where the number of images per class is small (on the order of hundreds). On CIFAR-10, however, where the number of image per class is an order of magnitude larger, cutout shows significant improvement over M\"obius. 
%     \item M\"obius with cutout significantly outperforms either augmentation on CIFAR-100, suggesting that the two techniques are complementary. This is important, as we designed M\"obius to combine easily with other augmentation methods.
% \end{itemize}

As shown in Table~\ref{tab:main}, these experiments highlight several key observations:
\begin{itemize}
    \item Empirically, M\"obius is able to achieve a higher accuracy than cutout on average.
    \item M\"obius with cutout significantly outperforms all other conditions on CIFAR-100 and Tiny ImageNet, suggesting that the two techniques are complementary. This is important, as we designed M\"obius to combine easily with other augmentation methods.
    \item In particular,  this improvement is apparent where the number of images per class is small (on the order of hundreds). On CIFAR-10, however, where the number of images per class is an order of magnitude larger, cutout shows significant improvement over M\"obius. 
\end{itemize}

% As shown in Table~\ref{tab:main}, these experiments highlight that the effectiveness of M\"obius on low data settings. In particular, M\"obius and M\"obius with cutout demonstrate significant improvements over cutout alone or standard crop-and-flip on Tiny ImageNet and CIFAR-100, respectively, where the number of images per class is small (on the order of hundreds). On CIFAR-10, however, where the number of image per class is an order of magnitude larger, cutout shows significant improvement over M\"obius. Finally, M\"obius with cutout significantly outperforms either augmentation on CIFAR-100, suggesting that the two techniques can be complementary. This is important, as we designed M\"obius to combine easily with other augmentation methods.

The results of this experiment suggest that M\"obius data augmentation can improve over cutout, the state-of-the-art performance on sample-level and label-preserving augmentation strategy. This effect is especially prominent in low data regimes, where there are fewer (on the order of hundreds) of samples per class. Provided that the distortions generated by M\"obius are highly varied, we expect, and observe empirically, that M\"obius data augmentation performs significantly better on the larger image resolutions of Tiny ImageNet over CIFAR10 or CIFAR100. The increased number of pixels permits a greater diversity of available $\mathcal{M}$-admissible M\"obius transformations. 
% , particularly when combined as complementary methods

% \def\checkmark{\tikz\fill[scale=0.4](0,.35) -- (.25,0) -- (1,.7) -- (.25,.15) -- cycle;} 
\newcommand{\subfigwidthtiny}{0.09}

\newcolumntype{P}[1]{>{\centering\arraybackslash}p{#1}}

\begin{table*}[t]
\centering
% \caption{Analysis of Animate and Inanimate Superclasses}
\begin{tabular}{P{2.6cm}P{1.1cm}P{1.1cm}P{.3cm}|P{2.7cm}P{1.1cm}P{1.1cm}P{.3cm}} 
\Xhline{3\arrayrulewidth} 
\hline
{\bf Animate} & {\bf Baseline} & {\bf  M\"obius} & {\centering\contour{black}{$\uparrow$}} & {\bf Inanimate} & {\bf Baseline} & {\bf M\"obius} & {\centering\contour{black}{$\uparrow$}} \\
\Xhline{3\arrayrulewidth}
Aquatic Mammals                & 73.0\%    & 74.4\%  & \checkmark & Large Natural Outdoor Scenes & 85.2\%  & 87.2\%  & \checkmark \\\hline
Fish                           & 81.8\%   & 82.4\%  & \checkmark & Food Containers    & 79.6\%   & 80.6\%  & \checkmark  \\\hline
Insects                        & 85.0\%    & 85.8\%  & \checkmark & Fruit and Vegetables   & 86.2\%  & 89.6\%  & \checkmark \\\hline
Large Omnivores and Herbivores & 84.4\%   & 84.8\%  & \checkmark & Household Electrical Device & 86.4\%  & 87.0\% & \checkmark \\\hline
Large Carnivores               & 83.8\%   & 84.4\%  & \checkmark &  Vehicles 1    & 88.6\%   & 90.8\%  & \checkmark  \\\hline
Non-insect Invertebrates       & 81.4\%   & 82.0\%   & \checkmark &     Large Man-made Outdoor Things  & 89.8\%   & 89.6\%   \\\hline
Medium-sized mammals           & 82.6\%   & 85.2\% & \checkmark & Household Furniture   & 86.\%.   & 84.8\%   &   \\\hline
People                         & 64.6\%   & 66.8\%  & \checkmark &  Trees    & 75.8\%   & 74.4\%  &  \\\hline
Small Mammals                  & 74.8\%   & 79.0\%    & \checkmark &  Flowers  & 86.0\%    & 85.6\%  &    \\\hline
Reptiles                       & 73.8\%   & 77.2\%  &   \checkmark    &  Vehicles 2   & 91.8\%   & 90.2\%  &   \\\hline
\Xhline{5\arrayrulewidth}
Arthropod   & 73.65\%  & 76.26\% & \checkmark  & Artifact       & 66.10\%  & 66.51\% & \checkmark  \\\hline
Coelenterate & 77.60\%  & 78.00\% & \checkmark  & Geological Form            & 68.16\%  & 68.24\% & \checkmark \\\hline
Echinoderm  & 65.20\%  & 68.80\%    & \checkmark & Miscellaneous  & 69.08\%  & 69.57\% & \checkmark  \\\hline
Shellfish   & 60.80\%  & 65.60\% & \checkmark   & Natural Object & 74.40\%  & 75.12\%  & \checkmark  \\\hline
Vertebrate  & 72.64\%  & 72.79\% & \checkmark   &                                 &         &         &        \\
\Xhline{3\arrayrulewidth}
% \end{tabular}
% Miscellaneous  & 54.77\% & 58.77\% & \checkmark
% Natural Object  & 56.80\% & 58.80\% & \checkmark
% Geological Form & 57.60\% & 56.40\%  &  

\\
\multicolumn{8}{c}{
\includegraphics[width=\subfigwidthtiny\columnwidth]{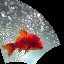} 
\includegraphics[width=\subfigwidthtiny\columnwidth]{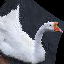} 
\includegraphics[width=\subfigwidthtiny\columnwidth]{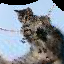}  \includegraphics[width=\subfigwidthtiny\columnwidth]{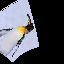} \hspace{4.5em} \includegraphics[width=\subfigwidthtiny\columnwidth]{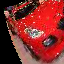} 
\includegraphics[width=\subfigwidthtiny\columnwidth]{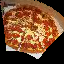} 
\includegraphics[width=\subfigwidthtiny\columnwidth]{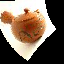} 
\includegraphics[width=\subfigwidthtiny\columnwidth]{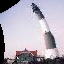}}
\end{tabular}
\\
\caption{Analysis of animate and inanimate superclasses, shown with animate \{goldfish, swan, cat, penguin\} and inanimate \{car, pizza, teapot, lighthouse\} image samples. M\"obius data augmentation consistently improves classification accuracy on animate superclasses (\checkmark on $\uparrow$), as opposed to inanimate superclasses on CIFAR (upper rows) and Tiny ImageNet (lower). On the Tiny ImageNet, M\"obius transformations improve performance on animate classes significantly more than inanimate classes. This empirical observation suggests that M\"obius transformations, having been studied in animals, are particularly attuned to improving generalization in these classes. Note that this finding is interesting, though by no means definitive or theoretically potent.}
\label{tab:animal}
\end{table*}

\subsection{Analysis on animate classes}
We analyze the predictions from M\"obius data augmentation by superclass in each dataset (Table~\ref{tab:animal}). Specifically, we compare two best model checkpoints trained on CIFAR-100: one with M\"obius and the other with the standard crop-and-flip baseline. In CIFAR-100, there are 20 superclasses with higher-level aggregations of five classes each~\cite{krizhevsky2009learning}. In our analysis, we find that the M\"obius-trained model improves performance on the 10 animate superclasses \{aquatic mammals, fish, insects, large omnivores and herbivores, large carnivores, non-insect invertebrates, medium-sized mammals, people, small mammals, reptiles\}. This contrasts with inconsistent performance differences among the inanimate superclasses. 

We perform a similar analysis by averaging the results of five baseline and M\"obius models on Tiny ImageNet, whose superclasses are derived from the ImageNet category tree.\footnote{http://www.image-net.org/api/xml/structure\_released.xml} We compare animate superclasses \{vertebrate, arthropod, coelenterate, shellfish, echinoderm\} with inanimate superclasses \{artifact, geological formation, natural object, miscellaneous\}. We find that M\"obius improves performance significantly more on animate classes than on inanimate classes compared to the standard baseline. 
% In fact, M\"obius improves significantly over the baseline on animate classes, while the improvement on inanimate classes is not significant. 
% Note that M\"obius improves overall performance over the standard baseline on all classes, so we draw on comparative improvements between the animate and inanimate classes.
% We find that M\"obius improves performance significantly more on animate classes ($2.31\%$) than on inanimate classes ($0.43\%$) compared to the standard baseline. 

% Note that M\"obius improves overall performance over the standard baseline on all classes.
% Specifically, M\"obius performs  better on average than the baseline on the animate classes, compared to  improvement on the inanimate classes on average. 
% The coelenterate superclass exhibits the same performance between the baseline and M\"obius.

These results suggest that M\"obius transformations, which have been studied in animals in prior literature~\cite{thompson1942growth, petukhov1989non, Lundh2011Cross-Development}, are especially effective on animate classes in image classification. While this finding is particularly interesting and consistent with scholarship, we heed that this observation remains empirical to this study and requires additional examination in order to be conclusive.

\section{Conclusion}
In this paper, we introduce M\"obius data augmentation, a method that applies $\mathcal{M}$-admissible M\"obius transformations to images during training to improve model generalization. Empirically, M\"obius performs best when applied to a small subset of data per class, e.g. in low data settings. M\"obius transformations are complementary to other sample-level augmentations that preserve labels, such as cutout or standard affine transformations. In fact, across experiments on CIFAR-10, CIFAR-100, and Tiny ImageNet, we find that cutout and M\"obius can be combined for superior performance over either alone. In future work, we plan to examine integrating them into the many successful data augmentation policies for fully supervised and semi-supervised learning. Ultimately, this work presents the first foray into successfully employing M\"obius transformations---the next level of mathematical abstraction from affine transformations---for data augmentation in neural networks and demonstrating the efficacy of this biologically motivated augmentation on image classification benchmarks.

% In future work, we plan to examine relaxing the constraints on possible M\"obius transformations for a more general construction as well as integrating them into the many successful data augmentation policies for fully supervised and semi-supervised learning.
% limitations:
% * no data aug schemes done (autoaugment, randaugment, uda - ssl in general, etc.)
% * did not compare with multi-sample techniques for data augmentation
% * mobius is constrained to 8 classes - future work to make this less constrained
% * future work is even more datasets to see how low we can go
\section*{Broader Impact}
Those who may benefit from this research largely include computer vision researchers and practitioners wishing to improve the generalization of their models without changing their model architectures or requiring additional large computational resources. This may specifically and especially benefit those who work on animal datasets, including the camera trap community who identify (rare) species in camera trap footage. As for biases, M\"obius has shown improvements on animate classes over inanimate ones. Consequently, this may result in imbalanced performance improvements, particularly on inanimate classes where this type of equivariance may not make sense to enforce.

\bibliography{refs}

\newcommand{\iu}{{i\mkern1mu}}
\onecolumn

\section*{Appendix A: Unconstrained M\"obius}

M\"obius data augmentation operates by constraining the group of M\"obius transformations that preserve image labels to the class of $\mathcal{M}$-admissible M\"obius transformations, discussed in greater detail in Section~\ref{sec:dataaug}. We run experiments comparing the performance of unconstrained M\"obius transformations in $\mathbb{R}^2$ against our method on all dataset settings. Figure~\ref{tab:random} displays a visual comparison. 

As shown in Table~\ref{tab:random} below, we observe that unconstrained M\"obius still outperforms the baseline of crop-and-flip transformations, even though it performs worse than our proposed method. Recall that the goal of the $\mathcal{M}$-admissible class is to prevent disruptive transformations, for example, that would cause only a single pixel to remain (similar to allowing ``crop" to crop the image down to 1 pixel). 

Our speculation is that most of the time, randomly parameterized M\"obius transformations are relevant to the image's invariance, which would improve the model's ability to generalize. Under fully unconstrained M\"obius, we would expect such transformations, that would unlikely improve regularization and may even hurt generalization, to occur more frequently.

% \newcolumntype{P}[1]{>{\centering\arraybackslash}p{#1}}

\begin{table}[h]
\caption{Juxtaposition of model performance using randomly parameterized M\"obius transformations with those using defined ones. M\"obius transformations with random parameters suffers in performance, though still better than the crop-and-flip baseline.}
\vspace{5mm}
\centering
\begin{tabular}{ccc}
\Xhline{2\arrayrulewidth}
\textbf{Augmentation Method} & \textbf{Dataset} & \textbf{Accuracy} \\
\hline
Crop-and-flip & C10   & 96.47\%   $\pm 0.04$    \\
M\"obius & C10   & \textbf{96.72\%}   $\pm \textbf{0.06}$    \\
Random M\"obius & C10   & 96.54\%   $\pm 0.06$  \\
\hline
Crop-and-flip & C100  & 81.91\%   $\pm 0.20$      \\
M\"obius & C100  & \textbf{82.85\%} $\pm \textbf{0.31}$      \\
Random M\"obius & C100  & 82.30\%   $\pm 0.11$  \\
\hline
Crop-and-flip & R C10  &  83.98\%    $\pm 0.16$     \\
M\"obius & R C10  &  \textbf{86.07\%}    $\pm \textbf{0.24}$ \\
Random M\"obius & R C10  &  85.58\%  $\pm 0.31$     \\
\Xhline{2\arrayrulewidth}
\end{tabular}
\label{tab:random}
\end{table}

\section*{Appendix B: Modulating the inclusion of M\"obius}
Given the inherent complexity of M\"obius transformations, we additionally explore the effects of incorporating an increasing amount of M\"obius transformations into the data augmentation process. We evaluate M\"obius representations of 10\% to 50\%, at increments of 10\% in between, on CIFAR-10 and CIFAR-100. The goal of this experiment is to examine the effects of modulating M\"obius representation during the training process. Note that the experiments in Section~\ref{sec:experiments-main} only focused on a stationary amount (20\%) of M\"obius. 

\begin{figure*}%
\centering
\textbf{Varying M\"obius representation in data augmentation} \\
\subfigure[CIFAR-10]{%
    \label{fig:graph10}%
    \includegraphics[width=.45\columnwidth]{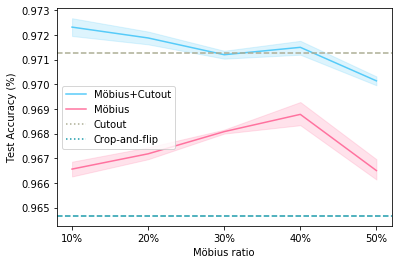}}%
\qquad
\subfigure[CIFAR-100]{%
    \label{fig:graph100}%
    \includegraphics[width=.45\columnwidth]{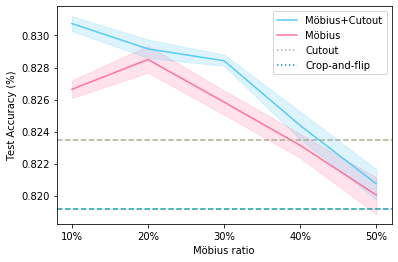}}%
\caption{Results from increasing M\"obius representation in data augmentation from 10\% to 50\% in 10\% increments, across 5 runs. \textbf{(a)} On CIFAR-10, M\"obius at only 10\% with cutout demonstrates empirically best results. M\"obius on its own performs best at 40\%, though it still performs under cutout alone. \textbf{(b)} On CIFAR-100, M\"obius reaches best performance at 20\% on its own and at 10\% with cutout. On both datasets, M\"obius boosts the performance of cutout when applied together, particularly in small quantities of 10-30\%.}
\label{fig:graph}
\end{figure*}

We compare these increments of M\"obius both with and without cutout. We then juxtapose these results with the baseline of cutout alone and that of standard crop-and-flip. We again report average performance and standard deviations across 5 runs on all experimental conditions. The results presented in Figure~\ref{fig:graph} emphasize the following findings:
\begin{itemize}
    \item Too much M\"obius data augmentation can result in disruptive training and poorer generalization. 
    \item M\"obius augmentation nevertheless outperforms both cutout and standard crop-and-flip baselines, across several values of representation particularly at 10\% and 20\% representation.
    % regardless of fractional inclusion between 10\% and 50\%, though at 50\% M\"obius starts to perform comparably to cutout.
    \item M\"obius augmentation alone experiences a local optimum at 40\% inclusion on CIFAR-10 and 20\% on CIFAR-100.
    \item M\"obius with cutout performs best with a very modest amount (10\%) of M\"obius. This is expected, as cutout provides additional regularization.
\end{itemize}

Though not shown in the graph, we also experiment with an even lower representation (5\%) of M\"obius in the M\"obius with cutout condition, in order to observe local optima and a bottoming out effect. We find that 10\% still shows superior performance to 5\% representation on both datasets. Specifically, M\"obius at 5\% with cutout performs 97.18\%$\pm 0.14$ on CIFAR-10 and 82.97\%$\pm 0.17$ on CIFAR-100. 

\section*{Appendix C: Defined M\"obius Parameters}
Given the inherent variability of M\"obius transformations, we additionally explore the effects of predefining a set of fixed M\"obius transformations as a way of decreasing variation and constraining the transformations to be human interpretable. Specifically, we define eight highly variable parameterizations that we visually verify to maintain their respective class labels. Based on their appearance, we describe them each as follows: (1) clockwise twist, (2) clockwise half-twist, (3) spread, (4) spread twist, (5) counter clockwise twist, (6) counter clockwise half-twist, (7) inverse, and (8) inverse spread. % Figure~\ref(fig:cifar}
Concretely, these parameters are presented below, where $\Re(p)$ and $\Im(p)$ denote the respective real and imaginary components of a point $p$, and height $x$ and width $y$ are dimensions of the original image.

Across all data settings, we found that the defined set of parameterizations performed better on average in experiments than our proposed class of $\mathcal{M}$-admissible M\"obius transformations, though the difference was not significant. This suggests that restraining variability to human interpretable transformations could improve model regularization and lead to improved generalization, though the difference is not significant. This is not extremely surprising, because M\"obius transformations can take on highly variable forms, for which some we may not expect or desire invariance. Nevertheless, this fixed method trades off the method's generalizability and ease of implementation.

Here is the precise parameterization:
\begin{enumerate}
    
    \item Clockwise twist: 
    \begin{fleqn}
    \begin{equation*}
     \begin{alignedat}{2}
        &\Re(z) = \{1, 0.5x, 0.6x\}, \\
        &\Im(z) = \{0.5y, 0.8y, 0.5y\}, \\
        &\Re(w) = \{.5x, 0.5x+0.3\sin(0.4\pi)y, 0.5x+0.1\cos(0.1\pi)y\}, \\
        &\Im(w) = \{y-1, 0.5y+0.3\cos(0.4\pi)y, 0.5y-0.1\sin(0.1\pi)x\}
    \end{alignedat}
    \end{equation*}
    \end{fleqn}
    
    \item Clockwise half-twist:
    \begin{fleqn}
    \begin{equation*}
     \begin{alignedat}{2}
        &\Re(z) = \{1, 0.5x, 0.6x\}, \\
        &\Im(z) = \{0.5y, 0.8y, 0.5y\}, \\
        &\Re(w) = \{.5x, 0.5x+0.4y, 0.5x\}, \\
        &\Im(w) = \{y-1, 0.5y, 0.5y-0.1x\}
    \end{alignedat}
    \end{equation*}
    \end{fleqn}

    \item Spread:
    \begin{fleqn}
    \begin{equation*}
     \begin{alignedat}{2}
        &\Re(z) = \{.3x, 0.5x, 0.7x\}, \\
        &\Im(z) = \{0.5y, 0.7y, 0.5y\}, \\
        &\Re(w) = \{0.2x, 0.5x, 0.8x\}, \\
        &\Im(w) = \{0.5y, 0.8y, 0.5y\}
    \end{alignedat}
    \end{equation*}
    \end{fleqn}
    
    \item Spread twist:
    \begin{fleqn}
    \begin{equation*}
     \begin{alignedat}{2}
        &\Re(z) = \{.3x, 0.6x, 0.7x\}, \\
        &\Im(z) = \{0.3y, 0.8y, 0.3y\}, \\
        &\Re(w) = \{0.2x, 0.6x, 0.8x\}, \\
        &\Im(w) = \{0.3y, 0.9y, 0.2y\}
    \end{alignedat}
    \end{equation*}
    \end{fleqn}
    
    \item Counter clockwise twist:
    \begin{fleqn}
    \begin{equation*}
     \begin{alignedat}{2}
        &\Re(z) = \{1, 0.5x, 0.6x\}, \\
        &\Im(z) = \{0.5y, 0.8y, 0.5y\}, \\
        &\Re(w) = \{0.5x, 0.5x+0.4y, 0.5x\}, \\
        &\Im(w) = \{y-1, 0.5y, 0.5y-0.1x\}
    \end{alignedat}
    \end{equation*}
    \end{fleqn}

    \item Counter clockwise half-twist:
    \begin{fleqn}
    \begin{equation*}
    \begin{alignedat}{2}
        &\Re(z) = \{1, 0.5x, 0.6x\}, \\
        &\Im(z) = \{0.5y, 0.8y, 0.5y\}, \\
        &\Re(w) = \{0.5x, 0.5x+0.3\sin(.4\pi)y, 0.5x+0.1\cos(.1\pi)x\}, \\
        &\Im(w) = \{y-1, 0.5y+0.3\cos(.4\pi)y, 0.5y-0.1\sin(.1\pi)x\}
    \end{alignedat}
    \end{equation*}
    \end{fleqn}
    
    \item Inverse:
    \begin{fleqn}
    \begin{equation*}
     \begin{alignedat}{2}
        &\Re(z) = \{1, 0.5x, x-1\}, \\
        &\Im(z) = \{0.5y, 0.9y, 0.5y\}, \\
        &\Re(w) = \{x-1, 0.5x, 1\}, \\
        &\Im(w) = \{0.5y, 0.1y, 0.5y\}
    \end{alignedat}
    \end{equation*}
    \end{fleqn}

    \item Inverse spread:
    \begin{fleqn}
    \begin{equation*}
     \begin{alignedat}{2}
        &\Re(z) = \{0.1x, 0.5x, 0.9x\}, \\
        &\Im(z) = \{0.5y, 0.8y, 0.5y\}, \\
        &\Re(w) = \{x-1, 0.5x, 1\}, \\
        &\Im(w) = \{0.5y, 0.1y, 0.5y\}
    \end{alignedat}
    \end{equation*}
    \end{fleqn}
    
\end{enumerate}

% CAN DISCUSS MORE HERE ON THIS
% We include additional experiments comparing the methods and further discussion on their comparison in the Appendix. We also release the code for predefined M\"obius alongside our original method at https://redacted.

% \newpage
\section*{Appendix D: M\"obius points mapping with and without interpolation}
We include visual representations of mapping M\"obius transformations from three points $\{w_1, w_2, w_3\}$ on the original image to three separate target points $\{z_1, z_2, z_3\}$ on the plane. In each example, the red, green, and blue points demonstrate various mappings between the two sets of three corresponding points. We also illustrate the effects of interpolation in filling in the gaps created by the M\"obius transformations.

{\centering
    \quad \textbf{Original} \quad \quad \textbf{Uninterpolated} \quad \textbf{Interpolated} \\
    \includegraphics[width=.45\columnwidth]{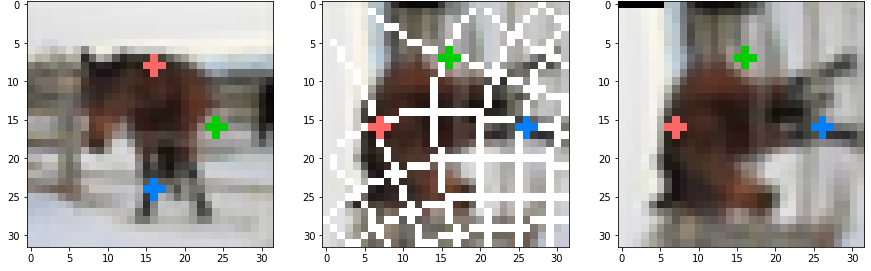} \\
    \includegraphics[width=.45\columnwidth]{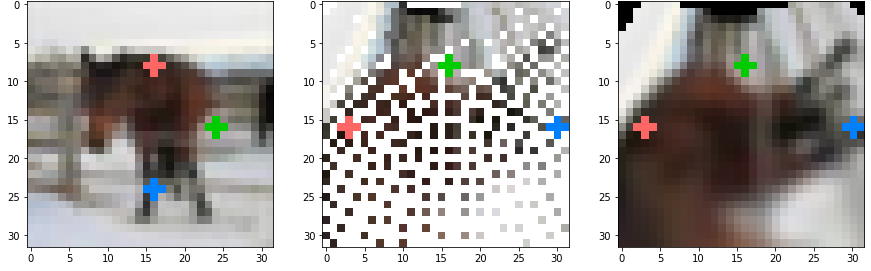} \\
    \includegraphics[width=.45\columnwidth]{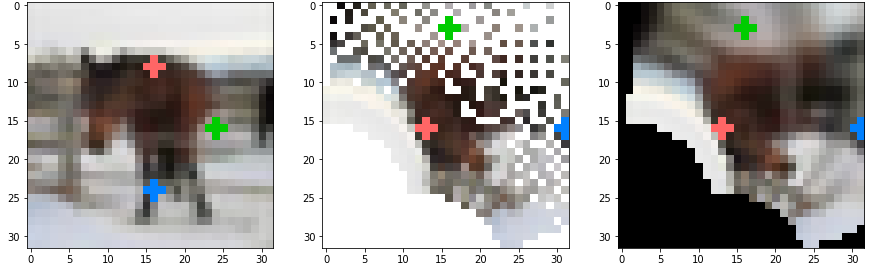} \\
    \includegraphics[width=.45\columnwidth]{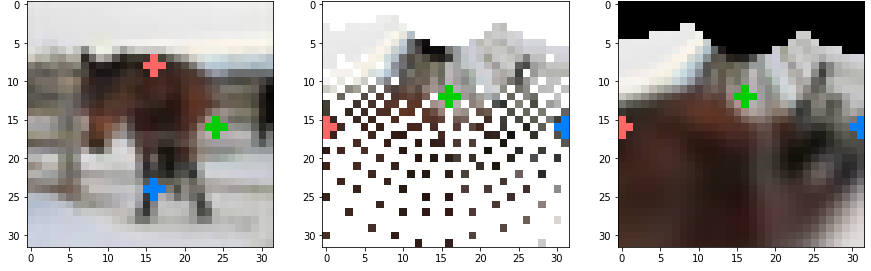} \\
    \includegraphics[width=.45\columnwidth]{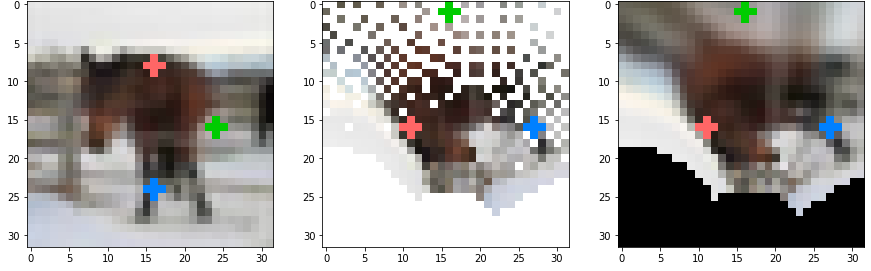} \\
}

Note that there is increased scatter of points when the mapped points are closer to the edge of the image and pixels are lost in the transformation, similar to scaling and cropping.

\end{document}